\def\year{2018}\relax
\documentclass[letterpaper]{article} 
\usepackage{aaai18}  
\usepackage{times}  
\usepackage{helvet}  
\usepackage{courier}  
\usepackage{url}  
\usepackage{graphicx}  
\usepackage{algorithm}
\usepackage{algpseudocode}

\usepackage{amsmath}
\usepackage{amsfonts}       
\usepackage{amssymb}
\usepackage{amsthm}
\usepackage{ifthen}

\usepackage{subfigure} 
\usepackage{graphicx} 

\begin{document}
%
\title{Learning Robust Options}
\author{Daniel J. Mankowitz$^1$, Timothy A. Mann$^{2}$, Pierre-Luc Bacon$^3$, Doina Precup$^3$, Shie Mannor$^1$ \\
$^1$ Technion Israel Institute of Technology, Haifa, Israel\\
$^2$ Google Deepmind, London, UK\\
$^3$ McGill University, Montreal, Canada\\
\small{danielm@campus.technion.ac.il, timothymann@google.com, pbacon, dprecup \{@cs.mcgill.ca\}, shie@ee.technion.ac.il}}
\maketitle

\newtheorem{assume}{Assumption}
\newtheorem{define}{Definition}
\newtheorem{lemma}{Lemma}
\newtheorem{theorem}{Theorem}
\newtheorem{corollary}{Corollary}

\newcommand{\Algorithm}{Robust Skills Policy Iteration}
\newcommand{\Alg}{IHOMP}
\newcommand{\TEA}{TEA}
\newcommand{\TEAs}{\TEA s}
\newcommand{\MProb}{MP}

\newcommand{\bellmanop}{\mathcal{T}}
\newcommand{\optionsNoInd}{\mathcal{O}}
\newcommand{\options}[1]{\optionsNoInd_{#1}}
\newcommand{\tri}[1]{TRI_{\rho_{#1}}}
\newcommand{\orho}[1]{o_{\rho_{#1}}}
\newcommand{\vmax}{V_{\rm MAX}}
\newcommand{\rmax}{R_{\rm MAX}}
\newcommand{\mset}{\{ 1, 2, \dots, m\}}
\newcommand{\shortmset}{[m]}
\newcommand{\grho}{\mathcal{G}_{\rho}}
\newcommand{\gib}{\mathcal{G}}

\newcommand{\xmark}{$\times$}

\begin{abstract}
Robust reinforcement learning aims to produce policies that have strong guarantees even in the face of environments/transition models whose parameters  have strong uncertainty. Existing work uses value-based methods and the usual primitive action setting.  In this paper, we propose robust methods for learning temporally abstract actions, in the framework of options. We present a Robust Options Policy Iteration (ROPI) algorithm with convergence guarantees, which learns options that are robust to model uncertainty. We utilize ROPI to learn robust options with the Robust Options Deep Q Network (RO-DQN) that solves multiple tasks and mitigates model misspecification due to model uncertainty. 
We present experimental results which suggest that policy iteration with linear features may have an inherent form of robustness when using coarse feature representations. In addition, we present experimental results which demonstrate that robustness helps policy iteration implemented on top of deep neural networks to generalize over a much broader range of dynamics than non-robust policy iteration.
\end{abstract}

%
%
\section{Introduction}
\label{sec:introduction}
In this paper, we focus on developing methods for learning temporally extended actions~\cite{Sutton1999} which are robust to model uncertainty. Temporally Extended Actions, also known as options \cite{Sutton1999}, skills \cite{daSilva2012,Mankowitz2016a,Mankowitz2016b} or macro-actions \cite{Hauskrecht1998} have been shown both theoretically \cite{Precup1998} and experimentally \cite{Mann2014a} to result in faster convergence rates in RL planning algorithms. We refer to a Temporally Extended Action as an option from here on in. While much research has been dedicated to automatically learning options, e.g. \cite{Simcsek2005,daSilva2012,Mankowitz2016a,Mankowitz2016b,Bacon2015},   no work has, to the best of our knowledge, focused on learning options that are robust to model uncertainty. 

\begin{figure*}[t!]
        \centering
        \includegraphics[width=1.0\textwidth]{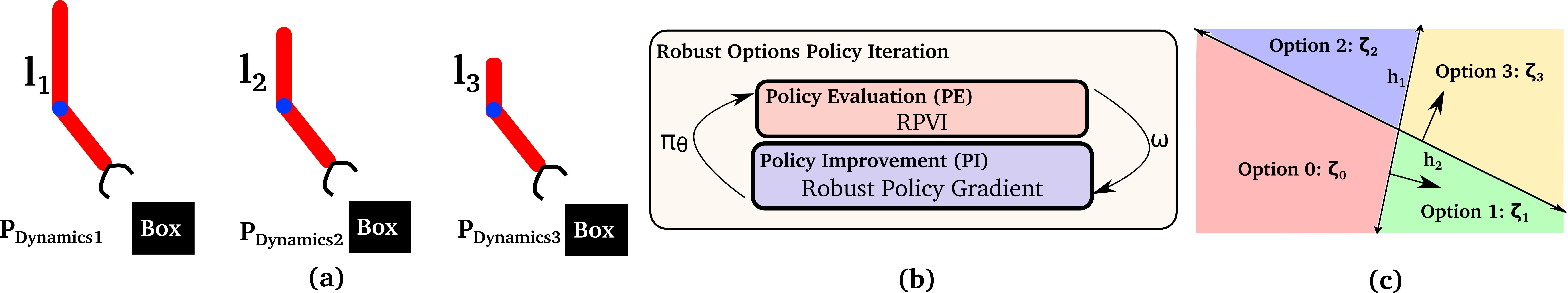}
   \caption{($a$) Parameter uncertainty in dynamical systems, also referred to as model uncertainty. An RL agent needs to be able to solve a given task
 for different parameter settings. ($b$) A high-level overview of the ROPI framework. ($c$) The option hyperplanes and option partitions for the ASAP option learning algorithm.}
   \label{fig:roadmap}
\end{figure*}

To understand model uncertainty, consider a two-link robotic arm that is trying to lift a box (Figure \ref{fig:roadmap}$a$). The arm with length $l_1$ can be modelled by a dynamical system $P_{dynamics1}$, also referred to as the state-transition function or transition model. These terms will be used interchangeably throughout the paper. The transition model governs the dynamics of this arm. Different models $P_{dynamics2}$ and $P_{dynamics3}$ are generated for arms with lengths $l_2$ and $l_3$ respectively. All of these arms are attempting to perform the same task. An RL agent trained using model $P_{dynamics1}$ may not adequately perform the task using $P_{dynamics2}$ or $P_{dynamics3}$. However, ideally the agent should be agnostic to the uncertainty in the  model parameters and still be able to solve the task (i.e., lift the box).  

Practical applications of RL rely on the following two-step blueprint: \textbf{Step one - Build a Model:} Models are attained in one of three ways - (1) A finite, noisy batch of data is acquired and a model is built based on this data; (2) A simplified, approximate model of the environment may be provided directly (e.g., power generation \footnote{\url{http://www.gridlabd.org/}}, mining etc); (3) A model of the environment is derived (e.g., dynamical systems). \textbf{Step two - Learn a policy:} RL methods are then applied to find a good policy based on this model. In cases (1) and (2), the parameters of the model are uncertain due to the noisy, finite data and the simplified model respectively. In case (3), 
model uncertainty occurs when the parameters of the physical agent are uncertain as discussed in the above-mentioned example. This is especially important for industrial robots that are periodically replaced with new robots that might not share the exact same physical specifications (and therefore have slightly different dynamical models). Learning a policy that is agnostic to the parameters of the model is crucial to being robust to model uncertainty. 

We focus on (3): Learning policies in \textit{dynamical systems}, using the robust MDP framework \cite{Bagnell2001,Nilim2005,Iyenger2005}, that are robust to model uncertainty (e.g., robots with different arm lengths). \footnote{Note that our theoretical framework can also deal with model uncertainty in cases (1) and (2).}


\textbf{Why learn robust options?} Previous works \cite{Mankowitz2016a,Mankowitz2016b,Bacon2015,Mankowitz2017} have shown that options mitigate Feature-based Model Misspecification (FMM). In the linear setting, FMM occurs when a learning agent is provided with a limited policy feature representation that is not rich enough to solve the task. In the non-linear (deep) setting, FMM occurs when a deep network learns a sub-optimal set of features, resulting in sub-optimal performance. We show in our work that options are indeed necessary to mitigate FMM. However, as discussed in the above example (Figure \ref{fig:roadmap}$a$) model uncertainty also results in sub-optimal performance. We show in our experiments that this is especially problematic in deep networks. Therefore, we learn robust options that mitigate both FMM and model uncertainty which we collectively term \textbf{model misspecification} \footnote{We do acknowledge that other forms of model misspecification exist. However, for this work we focus on FMM and model uncertainty.}.



Policy Iteration (PI) \cite{Sutton1998} is a powerful technique that is present in different variations \cite{Lagoudakis2003,Konda2000} in many RL algorithms. The Deep Q Network \cite{Mnih2015} is one example of a powerful non-linear function approximator that employs a form of PI. Actor-Critic Policy Gradient (AC-PG) \cite{Konda1999,Sutton2000} algorithms perform an online form of PI. As a result, we decided to perform option learning in a policy iteration framework. 

We introduce the Robust Options Policy Iteration (ROPI) algorithm that learns robust options to mitigate model misspecification, with convergence guarantees. Our novel ROPI algorithm consists of two steps, illustrated in Figure \ref{fig:roadmap}$b$, which include a Policy Evaluation (PE) step and a Policy Improvement (PI) step. For PE, we utilize RPVI \cite{Tamar2014} to perform policy evaluation and learn the value function parameters $w$; We then perform PI using the robust policy gradient (discussed in Section \ref{sec:rac-pg}). This process is repeated until convergence. ROPI learns robust options and a robust inter-option policy $\pi_{\theta}$, and has theoretical convergence guarantees. We showcase the algorithm in both linear and non-linear (deep) feature settings. 

In the linear setting we show that the \textit{non-robust} version of a linear option learning algorithm called ASAP \cite{Mankowitz2016b}, learns an inherently robust set of options. This, we claim, is due to the coarseness of the chosen feature representation. This provides evidence that, in some cases, linear approximate dynamic programming algorithms may get robustness `for free'. 

However, in the non-linear (deep) setting, explicitly incorporating robustness into the learning algorithm is crucial to being robust to model uncertainty. We incorporate ROPI into the Deep Q Network to form a Robust Options Deep Q Network (RO-DQN). Using the RO-DQN, the agent learns a robust set of options to solve multiple tasks (two dynamical systems) and mitigates model misspecification. 

\textbf{Main contributions: } (1) Learning robust options using our novel ROPI algorithm with convergence guarantees; This includes developing a Robust Policy Gradient (R-PG) framework, which includes a robust compatibility condition; (2) A linear version of ROPI that is able to mitigate model misspecification in Cartpole; (3) Experiments which suggest that linear approximate dynamic programming algorithms may get robustness `for free' by utilizing a coarse feature representation. (4) The RO-DQN, which solves multiple tasks by learning robust options, using ROPI, to mitigate model misspecification.

\section{Background}
\label{sec:background}
In this section, we relate the background material to the relevant module in the ROPI algorithm (PE or PI as shown in Figure \ref{fig:roadmap}$b$).


\textbf{Robust Markov Decision Process (PE):} A Robust Markov Decision Process (RMDP) \cite{Bagnell2001,Iyenger2005,Nilim2005} is represented by $\langle X,A,R,\gamma, P \rangle$ where $X$ is a finite set of states; $A$ is a finite set of actions, $R:X \times A \rightarrow \mathbb{R}$ is the immediate reward, which is bounded and deterministic, $\gamma \in [0,1]$ is the discount factor. Let $P: X \times A \rightarrow \mathcal{M}(X)$ be the transition function, mapping from a given state and action to a measure over next states. Given $x \in X$ and $a \in A$, nature is allowed to choose a transition to a new state from a family of transition probability functions $p \in \mathcal{P}(x,a)$. This family is called the uncertainty set \cite{Iyenger2005}. Uncertainty in the state transitions is therefore represented by $\mathcal{P}(x,a)$. 

The goal in a Robust MDP is to learn a robust policy $\pi:X \rightarrow A$ \cite{Iyenger2005}\footnote{In robust MDP literature the policy is often deterministic for notational convenience. A stochastic policy can be trivially derived by incorporating the uncertainty into the transitions}, which is a function mapping states to actions, that maximizes the \textit{worst case} performance of the robust value function ${V^{\pi}(x)=r(x,\pi(x))+\gamma\mbox{inf}_{p\in\mathcal{P}}\mathbb{E}^{p}[V^{\pi}(x')\vert x, \pi(x) ]}$ where $\mathcal{P}$ is the uncertainty set over state transitions; $r(x, \cdot)$ is the bounded immediate reward and $\mbox{inf}_{p\in\mathcal{P}}\mathbb{E}^{p}[V^{\pi}(x')\vert x,a]$ is the worst-case  expected return from state $x' \in X$ \cite{Iyenger2005,Nilim2005}. In order to solve this value function using policy evaluation, we define the robust operator ${\sigma_{\mathcal{P}(x,a)}:\mathbb{R}^{|X|}\rightarrow\mathbb{R}}$ for a given state $x$ and action $a$ where $\sigma_{\mathcal{P}(x,a)}v\doteq\mbox{inf}\{p^{T}v\,:\,p\in\mathcal{P}(x,a)\}$ and $v \in \mathbb{R}^{|X|}$ \cite{Iyenger2005}. They also defined the operator $\sigma_{\pi}:\mathbb{R}^{|X|}\rightarrow\mathbb{R}^{|X|}$ for a fixed policy $\pi$
such that $\sigma_{\pi}v(x)=\sigma_{\mathcal{P}(x,\pi(x))}v$. Using this operator, the robust value function is given by the following matrix equation $V^{\pi}=r+\gamma\sigma_{\pi}V^{\pi}$. 
It has been previously shown that the robust Bellman operator for a \textit{fixed policy} $T^{\pi}v\doteq r^{\pi}+\gamma\sigma_{\pi}v$ is a contraction in the sup-norm \cite{Iyenger2005} and the robust Bellman operator  $Tv(x)\doteq\mbox{sup}_{\pi}T^{\pi}v(x)$ is also a contraction with $V^{*}$ (the optimal value function for policy $\pi$) as its fixed point \cite{Iyenger2005}. 

\textbf{Robust Projected Value Iteration (PE):} Most of the Robust MDP literature has focused on small to medium-sized MDPs. \cite{Tamar2014} provide an approach capable of solving larger or continuous  MDPs using function approximation.  Suppose the value function is represented using linear function approximation \cite{Sutton1998}: $V(x) = \phi(x)^T w$, where $\phi(x) \in \mathbb{R}^d$ is a d-dimensional feature vector and $w \in \mathbb{R}^d$ is a set of parameters. Robust Projected Value Iteration (RPVI) \cite{Tamar2014}  involves solving the equation $\phi w_{k+1} = \Pi T^{\pi} (\phi w_k)$, where $\Pi$ is a projection operator onto the subspace spanned by $\phi$. Tamar et. al. show that $\Pi T^{\pi}$ is a contraction with respect to the sup-norm\footnote{A variation of this has also been shown for the average reward case \cite{Tewari2007}.}. This results in the update equation $w_{k+1}=(\Phi^{T}D\Phi)^{-1}(\Phi^{T}D\mathbf{r}+\gamma\Phi^{T}D\sigma_{\pi}\{\Phi w_{k}\})$ which can be sampled from data and solved efficiently for parameterized uncertainty sets that are convex in the parameters \cite{Tamar2014}. Here, $\Phi \in \mathbb{R}^{|X|\times d}$ is a matrix with linearly independent feature vectors in its rows and $D=diag(d)$ where $d$ is the state distribution for a policy $\pi$. RPVI utilizes a deterministic policy for notational convenience, but we assume a stochastic policy for the remainder of this paper. Note that this equation can be written as a robust critic update in Actor Critic Policy Gradient (AC-PG) algorithms, with a \textit{robust} TD error $\delta_k$, where the robust TD error is defined in Equation \ref{eqn:rtd}. The projection has been omitted since it can be viewed as a dynamic learning rate (see the Appendix for more details).

\begin{equation}\small
 \delta_k = r + \gamma \inf_{p \in \mathcal{P}(x,a)} \sum_{x'} p(x'\vert x,a) \phi(x')^Tw_k - \phi(x)^Tw_k \enspace .
 \label{eqn:rtd}
\end{equation} 
  



\textbf{Policy Gradient (PI):} Policy gradient is a standard technique in Reinforcement Learning that is used to estimate the parameters $\theta \in \mathbb{R}^d$ that maximize a performance objective $J(\pi_{\theta})$ via stochastic gradient descent \cite{Sutton1999}. A typical performance objective is the discounted expected return $J(\pi_{\theta})=\mathbb{E}[\sum_{t=1}^{\infty}\gamma^{t-1}r_t \vert x_0, \pi ]$ where $r_t$ is the reward at time $t$, $\gamma \in [0,1]$ is the discount factor and $x_0 \in X$ is a given start state. The gradient has been previously shown to be:

\begin{equation}
\frac{\partial J(\pi_{\theta})}{\partial \theta} = \sum_s d^{\pi}(x) \sum_a \frac{\partial \pi(x,a)}{\partial \theta} f_w(x,a) \enspace ,
\label{eqn:policy_gradient}
\end{equation}

where $d^{\pi}(x)$ is the discounted state distribution and $f_w(x,a)=\phi(x,a)^T w$ is an approximation of the action value function $Q^{\pi}(x,a)$. This gradient is then used to update the parameters ${\theta_{t+1} = \theta_t + \alpha_t \nabla_{\theta} J(\pi_{\theta})}$ for a stepsize $\alpha_t$.

\textbf{Options (PI)}: A Reinforcement Learning option \cite{Sutton1999,Konidaris2009} consists of a three-tuple $\zeta_i = \langle I, \xi_{\chi_{i}}, \beta(x) \rangle$, where $I$ is the set of initiation states from which the option can be executed; $\xi_{\chi_{i}}:X \rightarrow \Delta_A$ is the intra-option policy, parameterized by $\chi \in \mathbb{R}^d$ and is a mapping from states to a probability distribution over actions; $\beta(x)$ indicates the probability of the option terminating when in state $x\in X$.  

\textbf{Option Learning (PI):} There are some recent approaches to option learning but the approach we will focus on is the Adaptive Skills, Adaptive Partitions (ASAP) framework \cite{Mankowitz2016b}, which enables an agent to automatically learn both a hierarchical policy and a corresponding option set. The hierarchical policy learns where to execute options based on learning intersections of hyperplane half-spaces that divide up the state space. Figure \ref{fig:roadmap}$c$ contains an example of two option hyperplanes, $h_1$ and $h_2$, whose intersection divides the state space into four option partitions. Each option partition $i$ contains an option $\zeta_i$. The options and option partitions are learned using a policy gradient approach and are represented within the ASAP policy.
%
The ASAP policy $\pi:X \rightarrow \Delta_A$ is a function mapping a state to a probability distribution over actions and is defined as follows:

\begin{define}{(ASAP Policy).}
Given $K$ option hyperplanes, a set of $2^K$ options ${\Sigma=\{\zeta_{i} \vert i=1,\cdots 2^K\}}$, and an MDP $m$ from a space of  possible MDPs, the ASAP policy is defined as $\pi_{\chi, \beta}(a \vert x,m) = \sum_{i=1}^{2^K} p_{\beta}(i \vert x,m)\xi_{\chi_{i}}(a \vert x)$ where $p_{\beta}(i \vert x,m)$ is the probability of executing option $\zeta_i$ given that the agent is in state $x\in X$ and MDP $m \in M$; $\xi_{\chi_{i}}(a \vert x)$ represents the probability that, given option $i$ is executing, option $i$ will choose action $a \in A_{\zeta_i,x}$. 
\end{define}

\textbf{Deep Q Networks (PE+PI):} The DQN algorithm \cite{Mnih2015} is a powerful Deep RL algorithm that has been able to solve numerous tasks from Atari video games to Minecraft \cite{Tessler2016}. The DQN stores its experiences in an experience replay buffer \cite{Mnih2015} to remove data correlation issues. It learns by minimizing the Temporal Difference (TD) loss. Typically, a separate DQN is trained to solve each task. Other works have combined learning multiple tasks into a single network \cite{Rusu2015} but require pre-trained experts to train an agent in a supervised manner. Recently, robustness has been incorporated into a DQN \cite{Shashua2017}. However, no work has, to the best of our knowledge, incorporated \textit{robust} options into a DQN algorithm to solve multiple tasks in an online manner.

%

%
%

\section{Preliminaries}

Throughout this paper we make the following assumptions which are standard in Policy Gradient (PG) literature \cite{Bhatnagar2009}: \textit{Assumption (A1)}: Under any policy $\pi$, the Markov chain resulting from the Robust MDP is irreducible and aperiodic. \textit{Assumption (A2)}: A policy $\pi_\theta(x,a)$ for any $x\in X$, $a \in A$ pair is continuously differentiable with respect to the parameter $\theta$. In addition, we make \textit{Assumption (A3)}: The optimal value function $V^*$ is found within the hypothesis space of function approximators that are being utilized. We use these assumptions to define the robust transition probability function and the robust steady-state distribution for the discounted setting. We then use these definitions to derive the robust policy gradient version of Equation \ref{eqn:policy_gradient}. 


\subsection{Robust Transition Probability Distribution} 
The robust value function is defined for a policy $\pi:X\rightarrow A$ as ${V^{\pi}(x)=r(x,\pi(x))+\gamma\inf_{p\in\mathcal{P}}\mathbb{E}^{p}[V^{\pi}(x')\vert x,\pi ]}$ and the robust action value function is given by ${Q^{\pi}(x,a)=r(x,a)+\gamma\inf_{p\in\mathcal{P}}\mathbb{E}^{p}[V^{\pi}(x')\vert x,a,\pi]}$ where ${\hat{p}_{min}(x' \vert x,a)=\arg\inf_{p\in\mathcal{P}}\mathbb{E}^{p}[V^{\pi}(x')\vert x,a,\pi]}$. Here, $\hat{p}_{min}(x' \vert x,a)$ is the transition probability distribution that minimizes the expected return $\mathbb{E}^{p}[V^{\pi}(x')\vert x,a,\pi]$ for a given state $x$ and action $a$ and belongs to the pre-defined uncertainty set $\mathcal{P}$. Since $\hat{p}_{min}(x' \vert x,a)$ is selected independently for each state  we can construct a stochastic matrix $\hat{P}_{min}$ where each row is defined by $\pi^{\top}\hat{p}_{min}(x' \vert x,\cdot)$.

\subsection{Robust State Distribution} 
The matrix $\hat{P}_{min}$ can be interpreted as an adversarial distribution in a zero-sum game if the adversary fixes its worst case strategy \cite{Filar2012}. 


\begin{define}

Given the initial state distribution $\mu$, the robust discounted
state distribution is: ${\hat{d}^{\pi}(x)= \int\mu(x_{0})\sum_{t=0}^{\infty}\gamma^{t}\hat{P}_{min}^t(x|x_{0})dx_{0}}$.
\end{define}

The robust discounted state distribution is the same as the  state distribution used by \cite{Sutton2000} for the discounted setting. However, the transition kernel is selected robustly rather than assumed to be the transition kernel of the target MDP. The robust discounted state distribution $\hat{d}^{\pi}(x)$ intuitively represents executing the transition probability model that leads the agent to the worst (i.e., lowest value) areas of the state space. 

\section{Robust Policy Gradient}
\label{sec:rac-pg}
Using the above definitions, we can now derive the Robust Policy Gradient (R-PG) for the discounted setting, which is used for policy improvement in ROPI\footnote{Similar results are obtained for the average reward setting.}. To derive the R-PG, we (1) define the robust performance objective and (2) derive the corresponding robust compatibility conditions which enables us to incorporate a function approximator into the policy gradient.



\subsection{Robust Performance Objective} R-PG optimizes the discounted expected reward objective ${J(\pi)=\inf_{p \in \mathcal{P}} \mathbb{E}^{p} [\sum_{t=1}^\infty \gamma^{t-1}r_t \vert x_0, \pi, \mathcal{P} ]}$ where $\mathcal{P}$ is a given uncertainty set; $\gamma \in [0,1]$ is a discount factor and $r_t$ is a bounded reward at time $t$. Next we define the robust action value function as ${Q^{\pi}(x,a)=\inf_{p\in\mathcal{P}}\mathbb{E}^{p}[\sum_{t=1}^{\infty}\gamma^{t-1} r_{t}\vert x_{0}=x,a_{0}=a,\pi]}$ and we denote the robust state value function as $
{V^{\pi}(x)=\sum_{a\in A}\pi(x,a)Q^{\pi}(x,a)}$. The robust policy $\pi_{\theta}:X\rightarrow \Delta_A$ is parameterized
by the parameters $\theta\in\mathbb{R}^{d}$. We wish to maximize
$J(\pi_{\theta})$ to obtain the optimal set of policy parameters $\theta^{*}=\arg\max_{\theta}J(\pi_{\theta})$

\subsection{Robust Policy Gradient} Given the robust performance objective with respect to the robust discounted state distribution, we can now derive the robust policy gradient with respect to the robust policy $\pi_{\theta}$. $\pi$ is parameterized by $\theta$ unless otherwise stated. As in \cite{Sutton2000}, we derive the gradient using the robust formulation for the discounted scenario. The discounted expected reward case is presented as Lemma \ref{lemma:lemmagradavg}. The main differences between this Lemma and that of \cite{Sutton2000} is that we incorporate the robust state distribution $\hat{d}^{\pi}$ and emit a transition distribution $\hat{p}_{min}$ leading the agent to the areas of lowest value at each timestep.

\begin{lemma}
\label{lemma:lemmagradavg}
Suppose that we are maximizing the robust performance objective ${J(\pi)=\inf_{p\in\mathcal{P}}\mathbb{E}^{p}[\sum_{t=1}^{\infty}\gamma^{t-1}r_{t}\vert x_{0},\pi]}$
from a given start state $x_{0}\in X$ with respect to a policy ${\pi_{\theta}:X\rightarrow\Delta_{A}}$
parameterized by $\theta\in\mathbb{R}^{d}$ and the robust action
value function is defined as ${Q^{\pi}(x,a)=\inf_{p\in\mathcal{P}}\mathbb{E}^{p}[\sum_{t=1}^{\infty}\gamma^{t-1}r_{t}\vert x_{t}=x,a_{t}=a,\pi]}$.
Then the gradient with respect to the performance objective is:

\begin{equation}
\frac{\partial J(\pi)}{\partial\theta}= \sum_{x}\hat{d}^{\pi}(x)\sum_{a}\frac{\partial\pi(x,a)}{\partial\theta} Q^{\pi}(x,a) \enspace .
\end{equation}

\end{lemma}

The vectorized robust gradient update is therefore 
$\nabla_{\theta}J(\pi_{\theta})= \sum_{y}\hat{d}^{\pi}(x)\sum_{a}\nabla_{\theta}\pi_{\theta}(x,a)Q^{\pi}(x,a)$. It is trivial to incorporate a baseline into the lemma that does not bias the gradient leading to the gradient update equation: $\nabla_{\theta}J(\pi_{\theta})= \sum_{x}\hat{d}^{\pi}(x)\sum_{a}\nabla_{\theta}\pi_{\theta}(x,a)[Q^{\pi}(x,a)\pm b(x)]$.

\subsection{Robust Compatibility Conditions} The above robust gradient update does not as yet possess the ability to incorporate function approximation. However, by deriving robust compatibility conditions, we can replace $Q^{\pi}(x,a)$ with a linear function approximator $f_w(x,a)=w^T \psi_{x,a}$. Here $w\in \mathbb{R}^d$ represents the approximator's parameters and $\psi_{x,a}=\nabla\log \pi(x,a)$ which represent the compatibility features. The robust compatibility features are presented as Lemma \ref{lemma:lemmacompat}. Note that this compatibility condition is with respect to the robust state distribution $\hat{d}^{\pi}$.

\begin{lemma}
\label{lemma:lemmacompat}
Let $f_{w}:X\times A\rightarrow\mathbb{R}$ be an approximation to
$Q^{\pi}(x,a).$ If $f_{w}$ minimizes the mean squared error $e^{\pi}(w)=\sum_{x}\hat{d}^{\pi}(x)\sum_{a\in A}\pi(x,a)[Q^{\pi}(x,a)-f_{w}(x,a)]^{2}$
and is compatible such that it satisifes
$\frac{\partial f_{w}(x,a)}{\partial w}=\frac{\partial\pi(x,a)}{\partial\theta}\frac{1}{\pi(x,a)}$, then
\begin{equation}
\frac{\partial J(\theta)}{\partial\theta}= \sum_{x}\hat{d}^{\pi}(x)\sum_{a}\frac{\partial\pi(x,a)}{\partial\theta}f_{w}(x,a) \enspace .
\end{equation}
\end{lemma}

\section{Robust Options Policy Iteration (ROPI)}
\label{sec:rspi}


Given the robust policy gradient, we now present ROPI defined as Algorithm \ref{alg:my_alg}. A parameterized uncertainty set $p_\mu$ with parameters $\mu$ and a nominal model without uncertainty $\hat{p}_\mu$ are provided as input to ROPI. In practice, the uncertainty set, for example, can be confidence intervals specifying plausible values for the mean of a normal distribution. The nominal model can be the same normal distribution without the confidence intervals. At each iteration, trajectories are generated (Line 3) using the nominal model $\hat{p}_\mu$ and the current option policy $\pi_{\theta}$. These trajectories are utilized to learn the critic parameters $w$ in line $4$ using RPVI \cite{Tamar2014}.  As stated in the background section, RPVI converges to a fixed point. Once it has converged to this fixed point, we then use this critic to learn the option policy parameters $\theta$ (Line 5) using the R-PG update. This is the policy improvement step. This process is repeated until convergence. The convergence theorem is presented as Theorem \ref{thm:2}.   

%

\begin{algorithm}[tb] 

\caption{ROPI} 

\begin{algorithmic}[1] 
\Require ~\\ 
$w \in \mathbb{R}^d$ - The approximate value function parameters, 
$\phi \in \mathbb{R}^m$  - A set of features, $\pi_{\theta}$ - An arbitrary parameterized option policy with parameters $\theta$,  $\gamma$ - The discount factor, $p_\mu$ - An uncertainty set, $\hat{p}_\mu$ - A nominal model
 
\Repeat
\State $\tau \sim (\pi_{\theta}, \hat{p}_\mu)$ \Comment{Generate trajectories}
\State \textbf{PE}: \small $w_{t} = (\Phi^{T}D\Phi)^{-1}(\Phi^{T}D\mathbf{r}+\gamma\Phi^{T}D \sigma_{\pi_{\theta}}\{\Phi w_{t-1}\})$ \normalsize \Comment{Perform RPVI}
\State \textbf{PI}: \small $\theta_{t+1} =\theta_{t}+\alpha_{t}\nabla_{\theta} J_{\theta, w_{t}}(\pi_{\theta})$ \normalsize \Comment{Update policy parameters}
\Until {Convergence}
\State  {\bfseries return:}  {$\pi_{\theta}$} 
\Comment{ The robust option policy} 

\end{algorithmic} 
\label{alg:my_alg}

\end{algorithm}

\begin{theorem}
Let $\pi$ and $f_{w}$ be any differentiable function approximators
for the option policy and value function respectively that satisfy the compatibility
condition derived above and for which the option policy is differentiable
up to the second derivative with respect to $\theta.$ That is, $\max_{\theta,x,a,i,j}\vert\frac{\partial^{2}\pi(x,a)}{\partial\theta_{i}\partial\theta_{j}}\vert<B<\infty.$
Define $\{\alpha_{k}\}_{k=0}^{\infty}$ be any step-size sequence
satisfying $\lim_{k\rightarrow\infty}\alpha_{k}=0$ and $\sum_{k}\alpha_{k}=\infty$.
Then the sequence $\{J(\pi_{k})\}_{k=0}^{\infty}$ defined by any
$\theta_{0}$, $\pi_{k}(\cdot,\cdot,\theta_{k})$ and 

\begin{align*}
w_{k} & =w\enspace s.t. \enspace\sum_{x}\hat{d}^{\pi_{k}}(x)\sum_{a\in A}\frac{\partial\pi_{k}(x,a)}{\partial\theta} \\
 & \biggl[Q^{\pi_{k}}(x,a)-f_{w}(x,a) \biggr]\frac{\partial f_{w}(x,a)}{\partial w}=0\\
\theta_{k+1} & =\theta_{k}+\alpha_{k}\sum_{x}\hat{d}^{\pi_{k}}(x)\sum_{a}\frac{\partial\pi_{k}(x,a)}{\partial\theta}f_{w_{k}}(x,a)\enspace,
\end{align*}

converges such that $\lim_{k\rightarrow\infty}\frac{\partial J(\pi_{k})}{\partial\theta}=0$.
\label{thm:2}
\end{theorem}

%
%
\section{Experiments}
\label{sec:experiments}


We performed the experiments in two, well-known continuous domains called \textit{CartPole} and \textit{Acrobot} \footnote{\url{https://gym.openai.com/}}. The transition dynamics (models) of both Cartpole and Acrobot can be modelled as dynamical systems. For each experiment, the agent is faced with model misspecification. That is, \textit{Feature-based Model Misspecification (FMM)} and \textit{model uncertainty}. In each experiment, the agent mitigates FMM by utilizing options \cite{Mankowitz2016a,Mankowitz2016b,Mann2014b} and model uncertainty by learning robust options using ROPI. We analyze the performance of ROPI in the linear and non-linear feature settings. In the linear setting, we apply ROPI to the Adaptive Skills, Adaptive Partitions (ASAP) \cite{Mankowitz2016b} option learning framework. In the non-linear (deep) setting, we apply ROPI to our Robust Options DQN (RO-DQN) Network. 

The experiments are divided into two parts. In Section \ref{sec:linear}, we show that ROPI is not necessary as the learned linear `non-robust' options for solving CartPole provide a natural form of robustness and mitigate model misspecification. This provides some evidence that linear approximate dynamic programming algorithms which use coarse feature representations may, in some cases, get robustness `for free'. The question we then ask is whether this natural form of robustness is present in the deep setting? We show that this is \textbf{not} the case in our experiments in Section \ref{sec:nonlinear}. Here, robust options, learned using ROPI, are necessary to mitigate model misspecification. In each experiment, we compare (1) the misspecified agent (i.e., a policy that solves the task sub-optimally due to FMM and model uncertainty); (2) The `non-robust' option learning algorithm that mitigates FMM and (3) The robust option learning ROPI algorithm that mitigates FMM and model uncertainty (i.e., model misspecification). 

\subsection{Domains}

\subsubsection{Acrobot:} 
Acrobot is a planar two-link robotic arm in a vertical plane (working against gravity). The robotic arm contains an actuator at the elbow, but no actuator at the shoulder as shown in Figure \ref{fig:1}$a.1$. We focus on the \textit{swing-up} task whereby the agent needs to actuate the elbow actuator to generate a motion that causes the arm to swing up and reach the goal height shown in the figure. The state space is the $4$-tuple $\langle \theta_1, \dot{\theta_1}, \theta_2, \dot{\theta_2} \rangle$ which consists of the shoulder angle, shoulder angular velocity, elbow angle and elbow angular velocity respectively. The action space consists of torques applied to the elbow of $-1$ or $+1$. Rewards of $-1$ are received while the agent has not reached the goal, and $0$ received upon reaching the goal. The episode length is $500$ timesteps.

\subsubsection{CartPole:}
The CartPole system involves balancing a pole on a cart in a vertical position as shown in Figure \ref{fig:1}$a.2$. This domain is modelled as a continuous state MDP. The continuous state space consists of the $4$-tuple $\langle x,\dot{x}, \theta, \dot{\theta} \rangle$ which represent the cart location, cart horizontal speed, pole angle with respect to the vertical and the pole speed respectively. The available set of actions are constant forces applied to the cart in either the left or right direction. The agent receives a reward of $+1$ for each timestep that the cart balances the pole within the goal region (in our case, $\pm12$ degrees from the central vertical line) as shown in the figure. If the agent terminates early, a reward of $0$ is received. The length of each episode is $200$ timesteps and therefore the maximum reward an agent receives is $200$ over the course of an episode.

\subsection{Uncertainty Sets} 
For each domain, we generated an uncertainty set $\mathcal{P}$. In Cartpole, the uncertainty set $\mathcal{P}_{cartpole}$ is generated by fixing a normal distribution over the length of the pole $l_{pole}$, and sampling $5$ lengths from this distribution in the range $0.5-5$ meters prior to training. Each sampled length is then substituted into the cartpole dynamics equations generating $5$ different transition functions. A robust update is performed by choosing the transition function from the uncertainty set that generates the worst case value.  In Acrobot, the uncertainty set $\mathcal{P}_{acrobot}$ is generated by fixing a normal distribution over the mass of the arm link $m_{arm}$ between the shoulder and the elbow. Five masses are sampled from this distribution from $1-5$ Kgs and generated the corresponding transition functions. 

\subsection{Nominal Model} 
During training, in both Cartpole and Acrobot, the agent transitions according to the \textit{nominal} transition model. In Cartpole, the nominal model corresponds to a pole length of $0.5$ meters. In Acrobot, the nominal model corresponds to an arm mass of $1$ Kg. During evaluation, the agent executes its learned policy on transition models with different parameter settings (i.e., systems with different arm lengths in Cartpole and different masses in Acrobot). 

\begin{figure}[t!]
        \centering
        \includegraphics[width=0.45\textwidth]{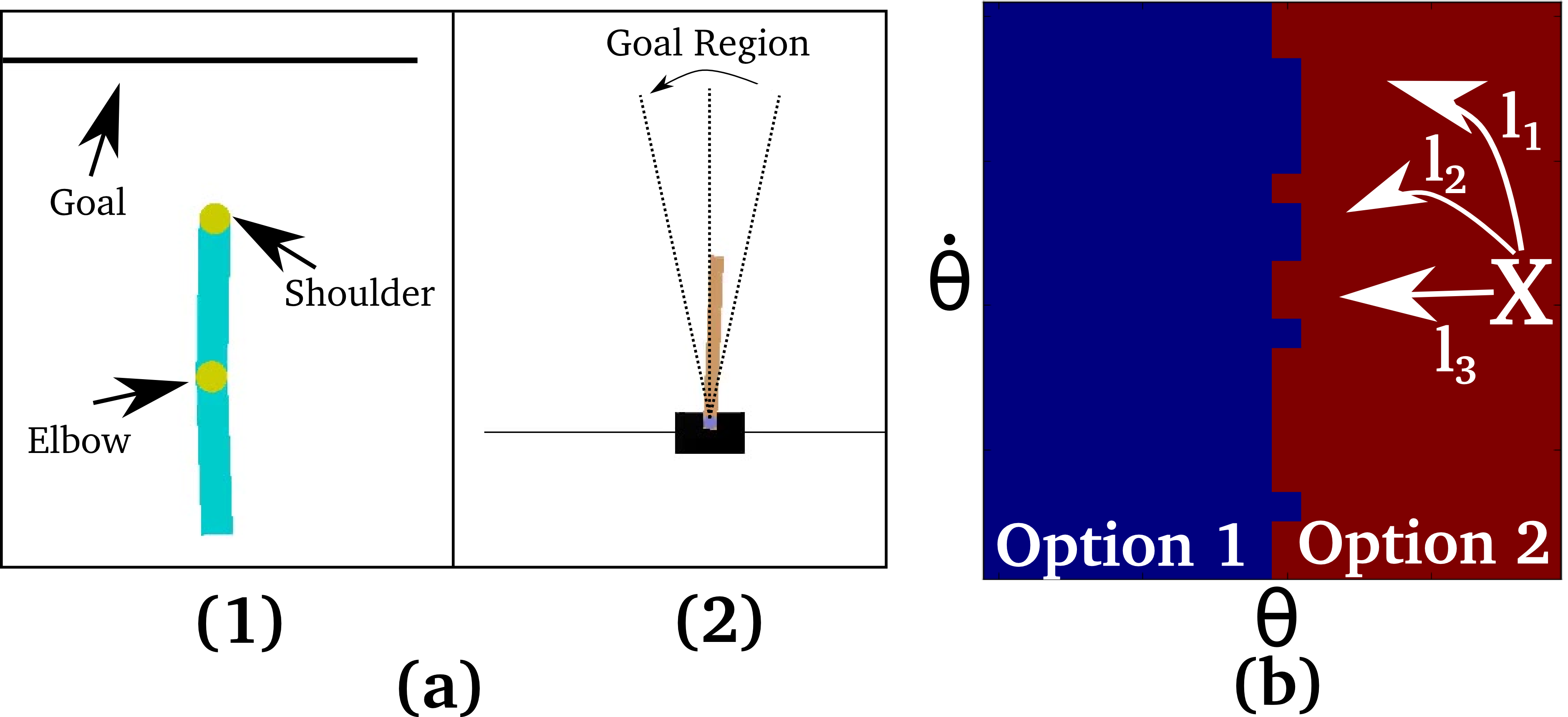}
   \caption{($a$)$.1$ The Cartpole and ($a$)$.2$ Acrobot domains. ($b$) Analysis of the option partitions in Cartpole.}
   \label{fig:1}
\end{figure}

\begin{figure*}[t!]
        \centering
        \includegraphics[width=1.0\textwidth]{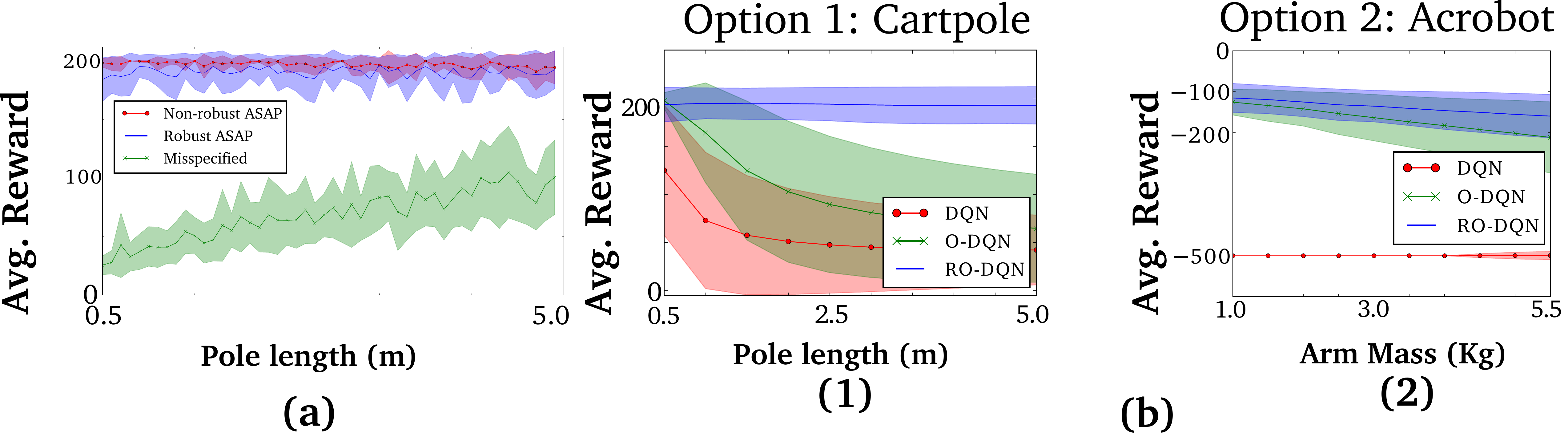}
   \caption{($a$) The learned options and option hyperplanes in the non-robust version of ASAP. ($b$) The average reward performance of the Robust Options DQN which learns options to solve ($b$)$.1$ CartPole and ($b$)$.2$ Acrobot. This is compared to the Option DQN and the misspecified agent (i.e., the regular DQN).}
   \label{fig:2}
\end{figure*}


\subsection{Linear ROPI: ASAP}
\label{sec:linear}
We first tested an online variation of ROPI on the Cartpole domain using linear features. To do so, we implemented a robust version of Actor Critic Policy Gradient (AC-PG) where the critic is updated using the robust TD error as shown in Equation \ref{eqn:rtd}. We used a constant learning rate which worked well in practice. The critic utilizes \textit{coarse} binary features which contain $[1,1,8,5]$ bins for each dimension respectively. We provide the actor with a limited policy representation, which is a probability distribution over the actions (left and right), independent of the state.  

We then trained the agent on the nominal pole length of $0.5$ meters. For evaluation, we averaged the performance of each learned policy over $100$ episodes per parameter setting, where the parameters were pole-lengths in the range $0.5-5.0$ meters.  As seen in Figure \ref{fig:2}$a$, the agent cannot solve the task using the limited policy representation, for any pole length, resulting in FMM. To mitigate the misspecification, we learn non-robust options using the ASAP algorithm \cite{Mankowitz2016b}. Using a single option hyperplane $K=1$ (see Section \ref{sec:background}), ASAP learns two options where each option's intra-option policy contains the \textit{same} limited policy representation as before. It is expected that the ASAP options mitigate the FMM and solve for pole lengths around the nominal pole length of $0.5$ meters on which it is trained. It should however struggle to solve the task for significantly different pole lengths (i.e., model uncertainty). 

To our surprise, the ASAP option learning algorithm was able to solve for all pole lengths over $0.5$ meters as shown in Figure \ref{fig:2}$a$, even though it was only trained on the nominal pole length of $0.5$ meters. Even after grid searches over all of the learning parameters, the agent still solved the task across these pole lengths. This is compared to a robust version of ASAP (Figure \ref{fig:2}$a$) that mitigated the misspecification and solved the task across multiple pole lengths as was expected.

We decided to analyze the learned `non-robust' options from the ASAP algorithm. Figure \ref{fig:1}$b$ shows the learned option hyperplane that separates the domain into two different options. The $x$ axis represents $\theta$ and the y axis $\dot{\theta}$. The red region indicates a learned option that always executes a force in the right direction. The blue region indicates a learned option that executes a force in the left direction. The learned option execution regions cover approximately half of the state space for each option. Therefore, if the agent is at point \textbf{X} in Figure \ref{fig:1}$b$, and the pole length varies (e.g., $l_1,l_2$ and $l_3$ in the figure), the transition dynamics will propagate the agent to slightly different regions in state space in each case. However, these transitions generally keep the agent in the correct option execution region, due to the coarseness of the option partitions, providing a natural form of robustness. This is an interesting observation since it provides evidence that linear approximate dynamic programming algorithms with coarse feature representations may, in some cases, get robustness `for free'. The question now is whether this natural form of robustness can be translated to the non-linear (deep) setting?


\subsection{Non-linear ROPI: RO-DQN}
\label{sec:nonlinear}
In the non-linear (deep) setting, we train an agent to learn robust options that mitigate model misspecification in a multi-task  scenario\footnote{In our setup, multi-task learning better illustrates the use-case of robust options mitigating model misspecification, compared to the single task setup where the use-case is less clear.}. Here, the learning agent needs to learn an option to solve Cartpole and an option to solve Acrobot using a common shared representation (i.e., a single network). The single network we use for each experiment is a DQN variant consisting of $3$ fully-connected hidden layers with $128$ weights per layer and ReLu activations. The hyper-parameter values can be found in the Appendix. We optimize the DQN loss function using the ADAM optimizer for a maximum of $3000$ episodes (unless the tasks are solved earlier). For evaluation, each learned network is averaged over $100$ episodes per parameter setting (i.e., parameter settings include pole lengths from $0.5-5.0$ meters for Cartpole and masses from $1.0-5.5$ Kgs for Acrobot).

In this setting, the DQN network struggles to learn good features to solve both tasks simultaneously using a common shared representation. It typically oscillates between sub-optimally solving each task, resulting in model misspecification\footnote{While different modifications can potentially be added to the DQN to improve performance \cite{Anschel2016,Van2016,Wang2015,Ioffe2015}, the goal of this work is to show that without these modifications, options can be used to mitigate the model misspecification.}. The average performance of the trained DQN on CartPole and Acrobot across different parameter settings is shown in Figures \ref{fig:2}$b.1$ and \ref{fig:2}$b.2$ respectively. 

We therefore add options to mitigate the model misspecification. The Option DQN (O-DQN) network utilizes two `option' heads by duplicating the last hidden layer. The training of these heads is performed in an alternating optimization fashion in an online manner (as opposed to policy distillation which uses experts to learn the heads with supervised learning \cite{Rusu2015}). That is, when executing an episode in Cartpole or Acrobot, the last hidden layer corresponding to Cartpole or Acrobot is activated respectively and backpropagation occurs with respect to the relevant option head. This network is able to learn options that solve both tasks as seen in Figures \ref{fig:2}$b.1$ and \ref{fig:2}$b.2$ for CartPole and Acrobot respectively. However, as the parameters of the tasks change (and therefore the transition dynamics), the option performance of the O-DQN in both domains degrades. This is especially severe in Cartpole as seen in the figure. Here, \textbf{robustness} is crucial to mitigating model misspecification due to uncertainty in the transition dynamics.

We incorporated robustness into the O-DQN to form the \textbf{Robust Option} DQN (RO-DQN) network. This network performs an online version of ROPI and is able to learn robust options to solve multiple tasks in an online manner. The main difference is that the DQN loss function now incorporates the robust TD update discussed in Section \ref{sec:background}. More specifically, the robust TD error is calculated as 
$${\delta = Q(x,a) - r + \gamma \inf_{p \in \mathcal{P}(x,a)} \sum_{x'}p(x' \vert x,a) \max_{a'}Q(x',a')}$$ \cite{Shashua2017}. The RO-DQN was able to learn options to solve both CartPole and Acrobot across the range of parameter settings as seen in Figures \ref{fig:2}$b.1$ and \ref{fig:2}$b.2$ respectively (see the Appendix for a combined average reward graph).


\section{Discussion}
\label{sec:discussion}
We have presented the ROPI framework that is able to learn options that are robust to uncertainty in the transition model dynamics. ROPI has convergence guarantees and requires deriving a Robust Policy Gradient and the corresponding robust compatibility conditions. This is the first work of its kind that has attempted to  learn robust options. In our experiments, we have shown that the linear options learned using the `non-robust' ASAP algorithm have a natural form of robustness when solving CartPole, due to the coarseness of the option execution regions. However, this does not translate to the deep setting. Here, robust options are \textit{crucial} to mitigating model uncertainty and therefore model misspecification. We utilized ROPI to learn our Robust Options DQN (RO-DQN). RO-DQN learned robust options to solve Acrobot and Cartpole for different parameter settings respectively. Robust options can be used to bridge the gap in  \textit{sim-to-real} robotic learning applications between robotic policies learned in simulations and the performance of the same policy when applied to the real robot. This framework also provides the  building blocks for incorporating robustness into continual learning applications \cite{Tessler2016,Ammar2015} which include robotics and autonomous driving.

\section{Acknowledgements}
This research was supported by the European Community’s Seventh Framework Programme (FP7/2007-2013) under grant agreement 306638 (SUPREL).

\bibliographystyle{aaai}
\bibliography{tmann}
\end{document}